\documentclass[
twocolumn,
]{ceurart}

\usepackage{todonotes}
\usepackage{enumitem}
\usepackage{graphicx}
\usepackage{subfig}

\usepackage{color}

\usepackage{listings}
\newcommand{\eg}{e.g., }

\newcommand{\ie}{i.e., }

\newcommand{\etal}{\textit{et al.}\ }
\newcommand{\figref}[1]{Fig.~\ref{#1}}    

\newcommand{\secref}[1]{Section~\ref{#1}}

\newcommand{\appref}[1]{Appendix~\ref{#1}} 

\begin{document}

\copyrightyear{2022}
\copyrightclause{Copyright for this paper by its authors.
  Use permitted under Creative Commons License Attribution 4.0
  International (CC BY 4.0).}

\conference{RecSys in HR'22: The 2nd Workshop on Recommender Systems for Human Resources, in conjunction with the 16th ACM Conference on Recommender Systems, September 18--23, 2022, Seattle, USA.}

\title{Design of Negative Sampling Strategies \\for Distantly Supervised Skill Extraction}

\author[1,2]{Jens-Joris~Decorte}[%
email=jensjoris@techwolf.ai,
url=https://www.techwolf.ai,
]
\cormark[1]
\address[1]{Ghent University -- imec,
  9052 Gent, Belgium}
\address[2]{TechWolf,
  9000 Gent, Belgium}

\author[2]{Jeroen~Van~Hautte}[%
email=jeroen@techwolf.ai,
url=https://www.techwolf.ai,
]

\author[1]{Johannes~Deleu}[%
email=johannes.deleu@ugent.be,
]

\author[1]{Chris~Develder}[%
email=chris.develder@ugent.be,
]

\author[1]{Thomas~Demeester}[%
email=thomas.demeester@ugent.be,
]

\cortext[1]{Corresponding author.}

\begin{abstract}
Skills play a central role in the job market and many human resources (HR) processes. 
In the wake of other digital experiences, today's online job market has candidates expecting to see the right opportunities based on their skill set.
Similarly, enterprises increasingly need to use data to guarantee that the skills within their workforce remain future-proof.
However, structured information about skills is often missing, and processes building on self- or manager-assessment have shown to struggle with issues around adoption, completeness, and freshness of the resulting data.
These challenges can be tackled using automated techniques for skill extraction.
Extracting skills is a highly challenging task, given the many thousands of possible skill labels mentioned either explicitly or merely described implicitly and the lack of finely annotated training corpora.
Previous work on skill extraction overly simplifies the task to an explicit entity detection task or builds on manually annotated training data that would be infeasible if applied to a complete vocabulary of skills.
We propose an end-to-end system for skill extraction, 
based on distant supervision through literal matching. We propose and evaluate several negative sampling strategies, tuned on a small validation dataset, to improve the generalization of skill extraction towards implicitly mentioned skills, despite the lack of such implicit skills in the distantly supervised data. 
We observe that using the ESCO taxonomy to select negative examples from related skills yields the
biggest improvements, and combining three different strategies in one model further increases the performance, up to 8 percentage points in RP@5.
We introduce a manually annotated evaluation benchmark for skill extraction based on the ESCO taxonomy, on which we validate our models. We release the benchmark dataset for research purposes to stimulate further research on the task.
\end{abstract}


\maketitle

\section{Introduction}

Skill extraction is an information extraction task that aims to identify all skills mentioned in a text. 
It is essential for many HR applications, such as resume screening and job recommendation systems. 
A comparative survey on skill extraction indicates that research interest has steadily grown over the last decade~\cite{khaouja2021survey}.
Traditionally, skill extraction has been approached as finding and disambiguating entities in texts. 
These methods typically rely on a named entity recognition (NER) component based on phrase-matching or a trained LSTM model~\cite{zhao2015skill, sayfullina2018learning, jia2018representation}.
However, skills are often present implicitly as longer sequences of words (which we refer to as spans) or full sentences rather than being mentioned explicitly: over 85\% of unique required skills in job ads have been reported never to be explicitly mentioned~\cite{bhola-etal-2020-retrieving}.
Very recently, the work titled SkillSpan has reformulated skill extraction as a more flexible span detection task~\cite{zhang2022skillspan}. 
The authors released a dataset of job postings with span annotations and trained SpanBERT-based models to detect skill spans as a sequence labeling task. 
In follow-up work, the authors developed a classification model to link such a span to the corresponding coarse-grained skill group in ESCO~\cite{zhang2022kompetencer}.
To overcome the difficulty of labeling these spans, the authors relied on weak supervision by automatically selecting labels based on the ESCO search API~\cite{zhang2022kompetencer}.
Another study manually annotated job ads with soft skills, which were consolidated into a released dataset called FIJO~\cite{beauchemin2022fijo}. However, instead of using an exhaustive list of soft skills, they only incorporated four broad labels to decrease the difficulty of the annotation.
The skill \emph{extraction} task can also be reduced to binary skill \emph{detection}, again reducing the challenge compared to fine-grained skill extraction~\cite{tamburri2020dataops}.
These works follow a more relaxed formulation of skill extraction, but they all suffer from the difficulty of annotating a fine-grained training dataset.

Some work avoids this labeling difficulty completely by using readily available labeled datasets. 
For example, in \cite{bhola-etal-2020-retrieving}, an eXtreme Multi-Label Classification (XMLC) model was trained based on a corpus of job ads with attached skills provided by an online job ads platform. 
However, the authors reported that for that corpus, at least 40\% of the vacancies missed 20\% of explicitly stated skills in their labels.
Recent work \cite{vermeerusing} successfully reconstructed the BERT-XLMC approach on Dutch vacancy texts using the Dutch RobBERT model \cite{delobelle2020robbert}. The training dataset used for this work is however based on the output of an existing commercial skill extraction solution. \\

We propose a new end-to-end approach to fine-grained skill extraction that does not rely on a large hand-labeled training corpus. 
Instead, we ease the requirements on the training data such that it can be automatically collected through distant supervision.
We cast the multi-label skill classification task into independent binary classification problems, with skills labeled on the sentence level, to encompass both explicit and implicit skill descriptions. 
To the best of our knowledge, our work is the first one to tackle fine-grained skill extraction in such a flexible distant supervision setup.
Our distant supervision training set contains few false positives, due to the literal matching of known skills, which is a task with low ambiguity. 
However, we expect many false negatives, for skills not literally mentioned. This is quantified in \secref{subsec:evaluation}.
We investigate to what extent the distantly supervised training set can be leveraged at maximum effectiveness to train a fine-grained skill extraction system. 
To that end, we design a number of negative sampling strategies that can be used to tune the extraction model training process on a small annotated development set, covering only a fraction of all potential skills (0.2\%, to be precise, in our experimental setting).
Finally, in order to stimulate research on automated skill extraction, and to facilitate the comparison of future models with our results, we release\footnote{\url{https://github.com/jensjorisdecorte/Skill-Extraction-benchmark}} our development and test data, which is constructed on top of the
``SkillSpan'' dataset \cite{zhang2022skillspan}, adding annotations with the ESCO~\cite{ESCO} skill labels.

\begin{figure*}[ht]
\centering
\includegraphics[width=0.85\textwidth]{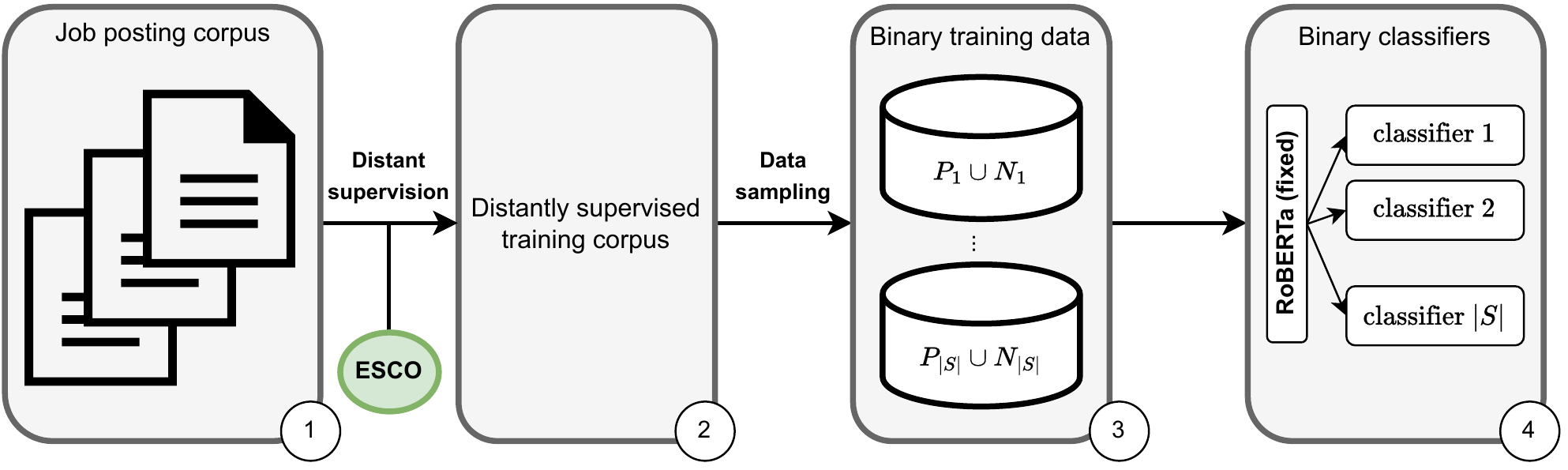}
\caption{Overview of our method. Using the ESCO skill taxonomy, the distantly supervised training corpus \raisebox{.5pt}{\textcircled{\raisebox{-.9pt} {2}}} is created from our job posting corpus \raisebox{.5pt}{\textcircled{\raisebox{-.9pt} {1}}}. Based on the negative sampling strategy, the positive data is combined with negative examples \raisebox{.5pt}{\textcircled{\raisebox{-.9pt} {3}}} and finally a classifier is trained for each skill \raisebox{.5pt}{\textcircled{\raisebox{-.9pt} {4}}}.} \label{fig:dataset}
\end{figure*}

\section{Related Work}

Multi-label classification datasets often have a skewed label distribution, with many labels occurring only a few times or even being completely absent in the training data. 
Some works have focused on improving the few-shot and zero-shot classification performance of multi-label text classification on these rare or unseen labels.
Typically, the information in structured label graphs (such as label descriptions or relations) or word embeddings are used as an input to the system in order to generalize to unseen labels~\cite{lu2020multi,chalkidis2019extreme}.
However, these methods still rely on a large labeled training dataset to work.
In the absence of any supervision, \cite{xiong2021extreme} uses a novel self-supervision training objective to train a dense sentence representation model that is used to assign labels based on cosine similarity in the learned space.
Yin \etal\cite{yin2019benchmarking} propose an entailment approach to zero-short text classification, where the input text is called the premise, and a hypothesis is constructed for each label using the template ``the text is about \emph{label}".
The premise and hypothesis are concatenated before being presented to a BERT-based model for prediction, making this method slow at inference for large label spaces. \\

Multi-label classification datasets not only suffer from the rare label problem, also many labels are just missing,
since they are usually only partially labeled:
instances without labels thus may either be truly negative, or positive but not identified as such during labeling.
The ``Single Positive Labels'' scenario is an extreme case of missing labels, where only one positive label is available for each training instance~\cite{cole2021multi}. 
Research on this topic is limited, and typically focuses on designing custom loss functions~\cite{zhou2022acknowledging} or online estimation of the missing labels during training~\cite{cole2021multi}.
This line of work is closely related to ``Positive-Unlabeled" (PU) binary classification, which is typically also tackled using custom loss functions~\cite{du2014analysis}.\\

Typically in a distant supervision setup, the labeling function is followed by a filtering step that aims to reduce the number of false positives in the labels~\cite{sterckx2016knowledge}.
However, we find that the number of false positives produced by the distant supervision step is low in our case of literal skill mentions.
This has been shown previously by \cite{decorte2021jobbert} where literal skill mentions have been successfully used as distant supervision for the task of job title representation learning.
Rather than focusing on a filtering step, we draw inspiration from the idea of ``hard negative examples'' in representation learning to improve the learning process.
In contrastive learning, hard negative examples refer to samples that are difficult to distinguish from an anchor point~\cite{robinson2020contrastive}.
This approach improves the discriminative abilities and downstream performance of unsupervised representation learning methods.
We adapt this idea to the multi-label classification setup, by oversampling negative examples from related labels.
More details on this approach are contained in the following section.

\section{Skill Extraction Approach}

We approach the task of skill extraction as a sentence-level multi-label classification task. 
A high-level overview of the method is shown in \figref{fig:dataset}.
Our method uses distant supervision based on the ESCO skill taxonomy to automatically assign (partial) skill labels for a given set of sentences from the HR domain (in particular, mined from vacancies).
Negative sampling strategies are used to combine `positive sentences' for a given skill (i.e., sentences labeled with that skill during the distant supervision step) with sentences not containing that skill (referred to as `negative sentences').
Finally, a binary classifier $f_s$ is trained for each skill $s$, based on the constructed positive and negative sentences for that skill.
It consists of a logistic regression classifier on top of a (frozen) representation for the sentences, as described in more detail below.

\paragraph{Distantly supervised training set:}
Given a set $S$ of skills and a background corpus of sentences $D$, for each skill $s \in S$, a set $P_s$ of positive sentences is collected from $D$ through distant supervision.
In particular, we use the ESCO~\cite{ESCO} skills taxonomy as the set of classification labels.
The set $P_s$ of positive sentences for each skill $s$, consists of those sentences in $D$ that literally mention the skill $s$ or any of its alternative forms, as provided in the taxonomy.
This assumes that there are no ambiguous skill names, which holds in most cases as skill names tend to be specific.
The positive labels are very precise, due to the distant supervision process based on literal matches with the highly specific ESCO skill names. However, this means potentially many skills remain unlabeled, i.e., the training data is prone to false negatives.
After the distant supervision step, on average 365 sentences were labeled per skill (for the set of 13,891 ESCO skills).
This dataset follows a long tail distribution, with 75.1\% of skills occurring in only ten or fewer sentences.

\paragraph{Skill extraction model:}
The model architecture is depicted in \figref{fig:dataset}.
We use a frozen pre-trained RoBERTa \cite{liu2019roberta} model with mean pooling to transform input sentences into fixed-length contextual representations, before presenting them for classification.
The classification is performed by separate binary text classification models $f_s$, each generating an independent prediction value for their respective skill label $s$.
In contrast to a typical multi-label model, we optimize each classification model separately on a different corresponding dataset, instead of training all weights together. 

\paragraph{Training with negative sampling:}
$P_s$ serves as positive training data for classifier $f_s$, and negative examples are sampled from the union of all positive sentence datasets of all other skills.
The basic mechanism for sampling negatives is uniform sampling from this union.
However, following the ideas in representation learning~\cite{robinson2020contrastive}, we hypothesize that sentences from related skills are more \textit{informative}, harder to distinguish from the positive sentences (i.e., closer to the decision boundary), and could thus improve the learning process. 
As such, a fraction of the negative examples are sampled specifically from sentences that are labeled with a related (but different) label to skill $s$.
We refer to these sentences as ``hard'' negative samples.
Our negative sampling strategy is thus defined by two important factors.
First, the fraction of uniformly sampled negatives versus the hard negative samples is important. 
Secondly, how we define whether two skills are related is crucial to the learning process. 
We introduce three different strategies for selecting the related skills in \secref{subsec:negativesampling}.

\paragraph{Inference and evaluation:}
The final model is used to rank the relevance of all skills for a given sentence. 
Similar to \cite{chalkidis2019extreme}, we use the macro-averaged R-Precision@K (RP@K) metric to evaluate the performance of the method.
Since predictions are made on a sentence-basis, we restrict the evaluation to low values of K. 
RP@K is defined in \eqref{eq:RP}, where the quantity $Rel(n,k)$ is a binary indicator of whether the $k^\text{th}$ ranked label is a correct label for data sample $n$, and $R_n$ is the number of gold labels for sample $n$.
In addition, we use the mean reciprocal rank (MRR) of the highest ranked correct label as an indicator of the ranking quality.
More information on the evaluation is presented in \secref{subsec:evaluation}.

\begin{equation}
    RP@K = \frac{1}{N} \sum\limits_{n=1}^{N} \sum\limits_{k=1}^{K} \frac{Rel(n,k)}{min(K, R_n)}
    \label{eq:RP}
\end{equation}

\begin {table*}[ht]
\begin{tabular*}{\textwidth}{@{\hskip 6pt\extracolsep{\stretch{1}}}*{4}{l}}
    \toprule
                        & Siblings                                  & Levenshtein                              & Embedding \\
    \midrule
disarm land mine        & ensure flock safety                       & find land mines                          & repair mine machinery \\
                        & protect important clients                 & search for land mines                    & handle mining plant waste \\
                        & signal for explosion                      & identify land mines                      & management of mine ventilation \\
                        & deal with challenging people              & dismantle machines                       & construct road base \\
    \midrule
Haskell                 & DevOps                                    & add smell                                & PostgreSQL \\
                        & XQuery                                    & upsell                                   & Erlang \\
                        & Windows Phone                             & sink wells                               & JavaScript \\
                        & SPARK                                     & speak well                               & C++ \\
    \midrule
manage musical staff    & discharge employees                       & manage musical groups                    & manage agricultural staff \\
                        & manage volunteers                         & manage musical events                    & manage staff \\
                        & supervise nursing staff                   & manage musicians                         & manage dental staff \\
                        & guide staff                               & manage educational staff                 & manage educational staff \\
    \bottomrule
\end{tabular*}
\caption{Examples of related skill labels for the three different selection strategies. The ``siblings'' examples are in no particular order as they form a set of siblings, rather than an ordered list.}
\label{table:negexamples}
\end{table*}

\subsection{Negative Sampling Strategies}
\label{subsec:negativesampling}

Rather than randomly sampling negative examples for training each binary skill classifier, we assume that sampling more \emph{informative} negatives will likely lead to a more efficient training procedure.
Instead of sampling hard negative sentences directly, we first identify related (yet different) skills, and then sample sentences with those labels.
We introduce three different strategies for identifying such related skills, which we analyze through the experiments defined in \secref{sec:experimentalsetup}.
The considered sets of related skills, given a particular skill $s$ are obtained as follows:
\begin{itemize}[leftmargin=*]
    \item \textbf{Siblings}: all skills that share a parent concept with $s$, as indicated by the ``broader concepts" field in ESCO.
    \item \textbf{Levenshtein}: The top 100 skills closest to $s$, according to their Levenshtein distance.
    \item \textbf{Embedding}: The top 100 skills closest to $s$ in terms of cosine similarity with their mean-pooled RoBERTa-encoded skill name representations.
\end{itemize}

For each of the negative sampling strategies, some example ESCO skills with their related labels according to the strategy are shown in table~\ref{table:negexamples}.

\section{Experimental setup}
\label{sec:experimentalsetup}

\subsection{Evaluation}
\label{subsec:evaluation}

While hand-labeling a training dataset for skill extraction is infeasible (given the huge number of skills, \eg over 13k in ESCO), we argue that with reasonable manual work, it is possible to construct a benchmark that can be used to compare the performance of different models.
We build upon the test set of the \emph{SkillSpan} dataset from~\cite{zhang2022skillspan}, which contains job posting sentences annotated with skill spans.
We manually annotate each span in SkillSpan with its corresponding ESCO skill (if it exists).
This span-based multi-class annotation is less complex than annotating complete sentences with multiple labels.
The process is performed on the test sets of the publicly released subsets \emph{TECH} and \emph{HOUSE}. 
Details on the annotation guidelines can be found in \appref{appendix:annotation}.
The annotation effort results in fine-grained ESCO skill labels for 64.5\% of the spans.
We split this dataset into a validation and test set using a 20\%/80\% split.
The validation set contains 165 unique skill labels, and over 80\% of the unique skill labels in the test set never occur in the validation set.
A more detailed breakdown of the number of spans and annotations is shown in table \tabref{table:dataset}.

\begin {table}[th]
\begin{tabular}{lcccccl}\toprule
& \multicolumn{2}{c}{\emph{TECH}} & \multicolumn{2}{c}{\emph{HOUSE}}
\\
\cmidrule(lr){2-3}\cmidrule(lr){4-5}
                                 & val    & test   & val   & test\\
\midrule
\emph{\# sentences}              & 470    & 1882   & 243   & 973 \\
\emph{\# spans}                  & 262    & 1024   & 191   & 786 \\
\midrule
\emph{\# spans with ESCO label}  & 152    & 644    & 131   & 532 \\
\bottomrule
\end{tabular}
\caption{Benchmark dataset statistics on the number of sentences, spans, and ESCO labeled spans, for both the \emph{TECH} and \emph{HOUSE} partitions of the SkillSpan dataset. Numbers of the validation and test split are indicated in the table.}
\label{table:dataset}
\end{table}

In order to verify our hypothesis that the distant supervision labeling leads to quite precise positive labels, at the cost of many false negatives, we validated the distant supervision labeling of the test set against the manual annotations.
The automatically assigned labels are indeed rather precise (overall precision of 79\%), but at the cost of low coverage (\ie a recall of 14.6\%).

\subsection{Experiments}
\label{subsec:experiments}

The sentences used for training are collected from a large proprietary corpus of public job postings.
This dataset has been collected from different public job boards and contains a large number of English job postings.
ESCO is used for the distant supervision step: a skill label is assigned when the skill itself, or one of its alternative forms provided by ESCO, is literally mentioned in a sentence.
For each skill classifier $f_s$, a maximum of one thousand positive sentences is retained.
The amount of negative examples per positive example is set to 10.
We train a baseline classifier without hard negative sampling.
In this case, all negative examples are sampled uniformly from the other positive corpora.
To investigate the optimal hard negative sampling procedure, we conduct a hyper-parameter search for the fraction of negatives sampled using the three strategies (sibling, levenshtein, embedding) versus uniform sampling. 
Based on the performance on the validation sets, we decide on an optimal value for this percentage.
Finally, we report the contribution of each of the negative sampling strategies when combined. 
This is reported based on performance on the unseen test set, and contributions of the strategies are shown through ablations, by leaving one strategy out at a time.
We refer to \appref{appendix:details} for more details on the training procedure.

\section{Results and Discussion}
\label{sec:results}

\begin{figure*}[b]
    \centering
    \subfloat[\centering \emph{TECH}]{{\includegraphics[width=0.47\textwidth]{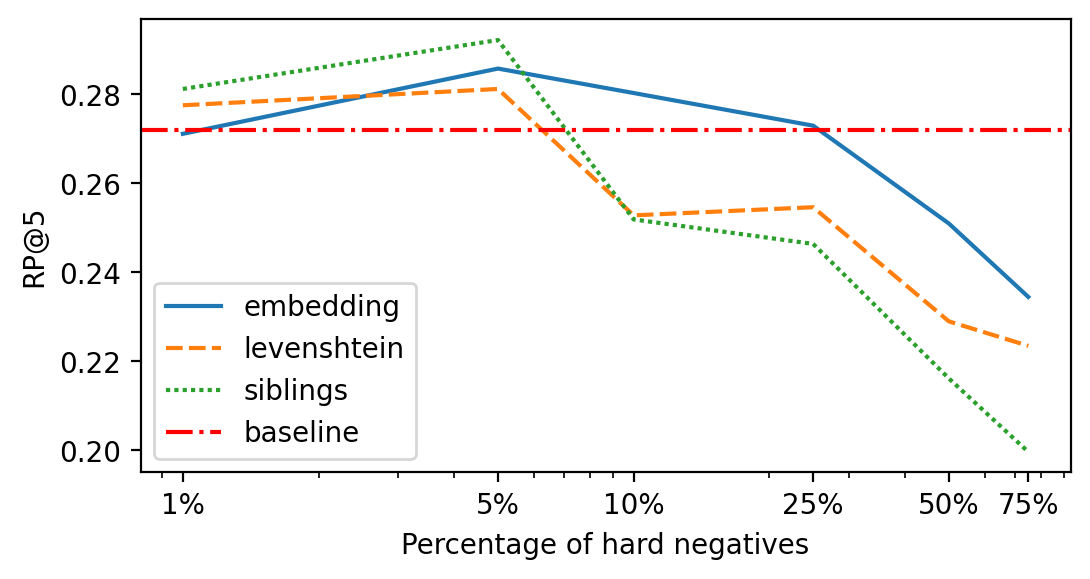} }}%
    \subfloat[\centering \emph{HOUSE}]{{\includegraphics[width=0.47\textwidth]{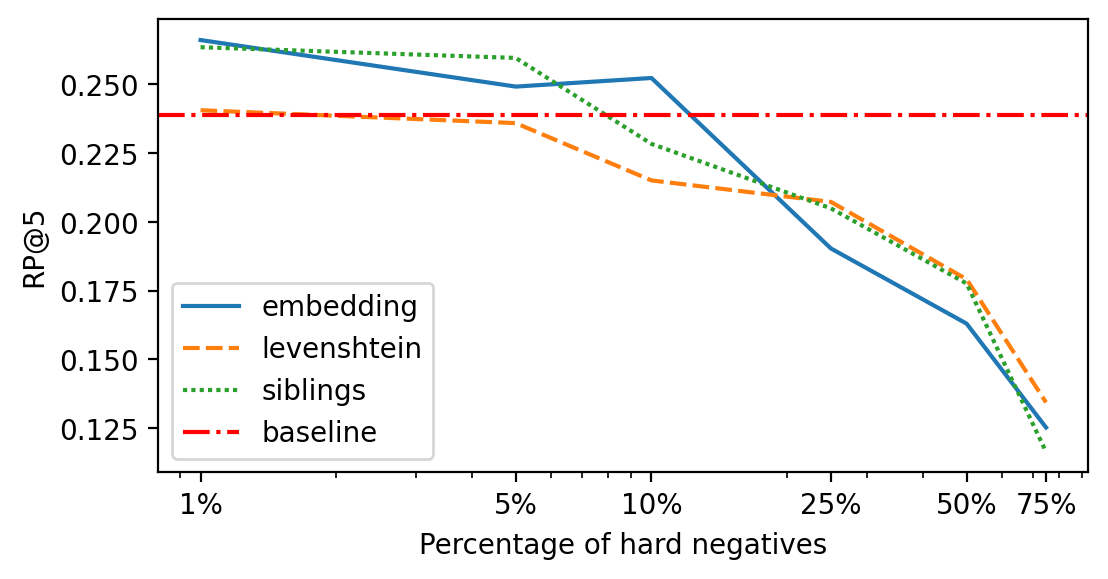} }}%
    \caption{Evaluating the effect of the fraction of hard negatives used during training, for each of the three strategies (siblings, levenshtein and embedding-based similarity with the considered positive skill) separately. The baseline model performance without hard negative sampling is shown by the horizontal red line. Metrics are reported on the validation sets.}%
    \label{fig:validation}%
\end{figure*}

The results of the hyper-parameter search for each of the negative sampling strategies are shown in \figref{fig:validation}. 
From these results, it is clear that the different strategies have different effects on the model performance.
Most notably, we find that the optimal fraction of hard negative sampling is no higher than 5\% for any strategy.
This is in line with previous findings on hard negative sampling~\cite{robinson2020contrastive}.
Sampling large amounts of hard negatives even has a large negative impact on the performance of the model.
Secondly, the ``levenshtein'' strategy brings the least improvements out of all three strategies.
\\

Finally, we trained a model that combines all strategies.
Based on the results of the above hyper-parameter search, we chose 5\% as an optimal value for the fraction of negatives sampled through the combined hard negative strategies.
To assess the impact of each of the strategies within this combination, we trained three more models in which each of the three strategies is left out respectively.
The performance of these final models is shown in table \tabref{table:results}.
The combination of all three strategies yielded the overall best model.
This model has large performance gains across the MRR and RP@K metrics for both the \emph{TECH} and \emph{HOUSE} dataset.
\\

Leaving out the ``Levenshtein'' strategy has a relatively low impact on the performance. 
This might be understood by looking at the examples in table \tabref{table:negexamples}: string similarity surfaces unrelated skills, for example for proper nouns such as \emph{Haskel}. 
This could partially explain the relatively low utility of this negative sampling strategy.
On the other hand, leaving out the ``siblings'' strategy takes away the largest part of the performance improvements.
This strategy makes use of the hierarchy defined in the ESCO taxonomy, and thus is a reliable method for selecting informative hard negatives.
The effect of the ``embedding'' strategy is comparable to the ``siblings'' strategy and thus proves a good alternative in case a hierarchy such as the one in ESCO is not available.

\begin {table*}[ht]
\begin{tabular}{lcccccl}\toprule
& \multicolumn{3}{c}{\emph{TECH}} & \multicolumn{3}{c}{\emph{HOUSE}}
\\\cmidrule(lr){2-4}\cmidrule(lr){5-7}
                                                           & MRR    & RP@5   & RP@10  & MRR    & RP@5   & RP@10\\\midrule
\emph{Baseline classifier}                                 & 0.246  & 23.65  & 33.71  & 0.255  & 26.66  & 34.19 \\\midrule
\emph{Classifier\textsubscript{neg}}                       & 0.326  & \textbf{31.71}  & 39.09  & \textbf{0.299}  & \textbf{30.82}  & \textbf{38.69} \\
\emph{Classifier\textsubscript{neg} without embeddings}    & 0.323  & 31.43  & \textbf{39.19}  & 0.298  & 29.09  & 37.70 \\
\emph{Classifier\textsubscript{neg} without Levenshtein}   & \textbf{0.339}  & 31.11  & 38.55  & 0.298  & 30.14  & 37.22  \\
\emph{Classifier\textsubscript{neg} without siblings}      & 0.303  & 30.57  & 37.07  & 0.281  & 29.20  & 35.91  \\
\bottomrule
\end{tabular}
\caption{Evaluation metrics of final skill extraction models on the \emph{TECH} and \emph{HOUSE} test sets. Reported metrics are mean reciprocal rank (MRR), R-Precision at 5 and at 10 (RP@5, RP@10).}
\label{table:results}
\end{table*}

\section{Conclusion and Future Work}

We propose an end-to-end approach to skill extraction using distant supervision. 
The method is able to make fine-grained skill predictions (using 13,891 skills from ESCO) for a given input sentence.
We introduce the idea of hard negative sampling through related labels in a multi-label classification setup and propose three different strategies to select these related labels.
We investigate the impact of each of the strategies, and found that all three strategies combined yield the highest increase on top of a baseline model without hard negative sampling.
Both the distant supervision and the hard negative sampling are designed to work well without manual labeling, which makes the whole method very flexible.
To the best of our knowledge, we are the first to design such a system for skill extraction, and we improve on prior work by providing methods that have relaxed the requirements from ground-truth data and that have the ability to make very fine-grained skill predictions.
Finally, we release our hand-labeled test and validation dataset for skill extraction to stimulate further research on the task.
\\

Future work could entail a more extensive investigation of other hyper-parameters, such as the number of negatives per positive sentence ($k$), which was fixed to 10 in this work. 
Secondly, more performance gains could be made if the RoBERTa weights were fine-tuned during training, but this requires changes in the training setup which should be carefully investigated.
Lastly, it could be interesting to investigate how limited manual labor can maximally improve the performance of the method even further with techniques such as active learning.

\begin{acknowledgments}
We thank the anonymous reviewers for their valuable feedback. This project was funded by the Flemish Government, through Flanders Innovation \& Entrepreneurship (VLAIO, project HBC.2020.2893).
\end{acknowledgments}

\bibliography{paper.bib}

\appendix

\section{Annotation guidelines}
\label{appendix:annotation}

Each item that needs to be annotated is a \textbf{span}, thus a part of a longer job posting sentence. Both the span and the complete sentence are shown to provide the right context for annotation. When a span is ambiguous, the full sentence must be read to understand the meaning of the span.\\

The task is to annotate the correct and most specific skill that is mentioned or implied by the span. The place of the candidate labels within the shortlist has no importance during annotation. In the case that no correct skill is found in the shortlist, you may search for the correct skill using the ESCO interface~\cite{ESCO}. If you still cannot find a correct label, select \emph{LABEL NOT PRESENT}. If you find that the span can generally not be interpreted as a skill, select \emph{UNDERSPECIFIED}.

\subsection{Examples}

\begin{itemize}[leftmargin=*]
\item Given the span ``partner continuously with your many stakeholders" and the candidate labels \emph{Communicate With Stakeholders}, \emph{Negotiate With Stakeholders} and ``Liaise With Shareholders", only the first two labels are considered correct. ``Communicate With Stakeholders" is most specific with regards to the span, so this label should be selected.
\item Spans such as ``apply your depth of knowledge" or ``apply your expertise" are classified as \emph{UNDERSPECIFIED}.
\end{itemize}

\section{Training details}
\label{appendix:details}

The separate classifiers are implemented as a simple logistic regression model, using the popular scikit-learn toolkit~\cite{scikit-learn}. All parameters are set to their default values, except for the inverse regularization strength parameter $C$, which is set to 0.1 for stronger regularization. The RoBERTa model and the mean pooling operation are implemented using the Sentence-BERT library~\cite{reimers2019sentence}.

\end{document}